\documentclass[11pt]{article}

\usepackage[preprint]{acl}
\usepackage{comment}

\usepackage{times}
\usepackage{latexsym}

\usepackage[T1]{fontenc}

\usepackage[utf8]{inputenc}

\usepackage{microtype}

\usepackage{inconsolata}

\usepackage{graphicx}
\usepackage{booktabs}
\usepackage{enumitem}
\usepackage{multirow}
\usepackage{tikz}
\usepackage{float}
\definecolor{adageblue}{HTML}{1565c0}
\definecolor{adagepeach}{HTML}{ef8c5e}
\usepackage{CJKutf8}
\usepackage[most]{tcolorbox}

\title{ADAGE: A Language-Agnostic Pipeline for Analogical Reasoning Evaluation}

\author{
 \textbf{Ahmed Haj Ahmed},
 \textbf{Alvin Grissom II}
\\
\\
Haverford College
\\
 \small{
   \textbf{Correspondence:} \href{mailto:ahajahmed@haverford.edu}{ahajahmed@haverford.edu}
 }
}

\begin{document}
\maketitle
\begin{abstract}
Multilingual reasoning evaluation overwhelmingly relies on translating English benchmarks, a practice that introduces linguistic artifacts and fails to test culturally-grounded reasoning. We introduce ADAGE (Analogical Difficulty-by-design Assessment for Grounded Evaluation), a language-agnostic pipeline that combines native-speaker curation with LLM-assisted generation to construct challenging, translation-free benchmarks for abstract analogical reasoning. We validate ADAGE by constructing benchmarks for Arabic, Amharic, and Japanese. Evaluating 14 open-weight models, we find a consistent cultural reasoning gap: models that perform well on English proverb reasoning struggle substantially on all three native benchmarks, with accuracy dropping by 12--52 percentage points relative to English. We release the pipeline, all three benchmarks, and the full evaluation suite.\footnote{Code and data: \url{https://github.com/AhmedHajAhmed/adage-benchmark}}
\end{abstract}

\section{Introduction}

Large language models have demonstrated strong performance on tasks commonly interpreted as requiring complex reasoning \citep{hendrycks2021measuringmassivemultitasklanguage, cobbe2021trainingverifierssolvemath, srivastava2023imitationgamequantifyingextrapolating, wei2023chainofthoughtpromptingelicitsreasoning}, yet this progress has been measured almost exclusively through English-centric benchmarks. The dominant approach to multilingual evaluation is to machine-translate these benchmarks into other languages \citep{fabbri2025multinrcchallengingnativemultilingual, ghosh2025surveymultilingualreasoninglanguage}---a practice which, we argue, creates the illusion of multilingual competence. Meanwhile, even in English, recent work has shown that benchmark performance can reflect brittle heuristics rather than robust reasoning \citep{mirzadeh2025gsmsymbolicunderstandinglimitationsmathematical}. If we cannot be confident of genuine reasoning in English, the situation is far worse when these already-fragile evaluations are simply translated.

Evaluation-by-translation is problematic for two reasons: first, machine translation introduces unnatural phrasing and subtle errors---translationese---that confound the measurement: a model's failure may reflect poor translation rather than poor reasoning \citep{graham2019translationesemachinetranslationevaluation, dey2024betteraskenglishevaluation}. Second, and more fundamentally, translation preserves the source culture's assumptions while discarding the target culture's reasoning patterns \citep{rystrøm2025multilingualmulticulturalevaluating}. Abstract analogical reasoning---interpreting a proverb like \textit{``He who does not know the falcon will roast it''} and applying it to a novel situation---cannot in general survive translation because the underlying cultural schema is non-transferable. Translation-based benchmarks therefore test whether models can apply English-centric analogies in another language, not whether they can reason \textit{within} another culture with different assumptions. 

\paragraph{Cultural Reasoning}
We use the term \textit{cultural reasoning} to refer narrowly to the appropriate application of abstract meaning using culturally shared metaphors, norms, and contextual assumptions not explicitly stated in text---a capacity that is intrinsically language- and culture-bound~\citep{GENTNER1983155, holyak_analogy}. 


Proverb interpretation is a prototypical instance of this capacity. Cognitive psychology has shown that proverb comprehension requires identifying structural correspondences between contexts that share little or no lexical overlap, rather than merely decoding literal meaning \citep{GENTNER1983155, proverb_comprehension_as_function}. Successful interpretation depends on executive functions and higher-order abstraction \citep{development_proverb_comprehension, holyak_analogy}: the reasoner must infer the underlying metaphor, map that relational structure to a novel context, and derive non-literal conclusions \citep{honeck1997proverb}. These processes are not purely linguistic but culturally situated---proverb comprehension arises from the interaction of metaphorical mapping, world knowledge, and cultural familiarity \citep{Alhazmi2025EmbodimentIC, liu-etal-2024-multilingual}. This makes proverb reasoning a stringent test of whether a system can recover abstract meaning and apply it flexibly, rather than relying on surface co-occurrence patterns.

A growing body of work has begun to address the gap between translation-based and native cultural reasoning through natively authored benchmarks. MultiNRC \citep{fabbri2025multinrcchallengingnativemultilingual} targets culturally grounded reasoning in French, Spanish, and Chinese; IOLBENCH \citep{goyal2025iolbenchbenchmarkingllmslinguistic} evaluates formal linguistic reasoning; and Jawaher \citep{magdy2025jawahermultidialectaldatasetarabic} provides a multidialectal Arabic proverb collection. ProverbEval \citep{azime2025proverbevalexploringllmevaluation} benchmarks proverb understanding in low-resource Ethiopian languages. However, these efforts share two limitations: each is constructed \textit{ad hoc} for a single language or language group, with no replicable pipeline that other communities can follow; and most test proverb explanation or meaning retrieval rather than analogical application: whether a model can map a proverb's abstract meaning onto a novel scenario. Recent evidence suggests that models can memorize proverbs yet fail to reason about them in culturally nuanced ways \citep{liu-etal-2024-multilingual}, underscoring the need for benchmarks that test application, not recall.

We introduce \textbf{ADAGE} (Analogical Difficulty-by-design Assessment for Grounded Evaluation), a language-agnostic, semi-synthetic pipeline for constructing culturally-grounded reasoning benchmarks without translation. ADAGE combines native-speaker curation with LLM-assisted scenario generation and a novel distractor strategy: proverbs are clustered by theme, and this structure is exploited to select same-cluster, cross-theme near-match, and random distractors, creating a controlled difficulty gradient within each item. Because every stage operates on abstract thematic structure rather than language-specific features, the pipeline transfers directly to any language with a corpus of proverbs or analogous cultural reasoning artifacts.

\paragraph{Contributions}
We validate ADAGE's portability by constructing benchmarks for three typologically diverse languages---Arabic, Amharic, and Japanese---spanning two language families, three scripts, and three resource levels. Evaluating 14 open-weight models (270M--35B parameters) from three families under zero-shot conditions, we find a consistent cultural reasoning gap: models that perform well on English proverb reasoning \citep{ghosh2022epicemployingproverbscontext} struggle substantially on all three native benchmarks, with accuracy dropping by 12--52 percentage points. Error analysis confirms that the distractor strategy functions as designed, with cross-theme near-matches attracting disproportionate errors whose gradient sharpens with model capability.

Our contributions are:

\begin{enumerate}
    \item \textbf{ADAGE}, a replicable, language-agnostic pipeline for constructing culturally grounded reasoning benchmarks, empirically validated across three typologically diverse languages.
    \item \textbf{Three benchmarks}---CAPR-ar, CAPR-am, and CAPR-ja (Culturally-grounded Analogical Proverb Reasoning)---for abstract analogical proverb reasoning in Arabic, Amharic, and Japanese, the first benchmarks targeting this task in these languages.
    \item \textbf{Empirical evidence} of a cultural reasoning gap that is consistent across model families and scales, demonstrating that translation-based evaluation overestimates multilingual reasoning capabilities.
\end{enumerate}

\section{Related Work}

\paragraph{Multilingual evaluation and its limits.}
Multilingual benchmarks have rapidly expanded language coverage. BenchMAX \citep{huang2025benchmaxcomprehensivemultilingualevaluation} evaluates across dozens of languages; MMATH \citep{luo2025mmathmultilingualbenchmarkmathematical} extends mathematical reasoning to multilingual settings; and XCOPA \citep{ponti2020xcopamultilingualdatasetcausal} tests causal commonsense across eleven languages. However, the dominant construction method remains translation from English, and the resulting benchmarks inherit English-centric reasoning structures. SynDARin \citep{ghazaryan2024syndarinsynthesisingdatasetsautomated} attempts to scale multilingual reasoning data by generating reasoning steps in English and translating them, perpetuating this dependency. Empirical evidence confirms the problem: models often perform better when prompted in English, even when the content is about other languages \citep{dey2024betteraskenglishevaluation, belay2025culemoculturallensesemotion}, suggesting that translated prompts are a degraded signal. More fundamentally, \citet{rystrøm2025multilingualmulticulturalevaluating} argue that multilingual fluency does not imply multicultural understanding; a model can generate grammatical Arabic while defaulting to Western assumptions about social norms and reasoning.

\paragraph{Native-authored and cultural reasoning benchmarks.}
A recent wave of work has moved beyond translation and toward natively-constructed evaluation. MultiNRC \citep{fabbri2025multinrcchallengingnativemultilingual} targets wordplay and culturally-grounded reasoning in French, Spanish, and Chinese. IOLBENCH \citep{goyal2025iolbenchbenchmarkingllmslinguistic} evaluates formal linguistic reasoning independent of cultural knowledge. CulturalBench \citep{chiu2025culturalbenchrobustdiversechallenging} and Global~PIQA \citep{chang2025globalpiqaevaluatingphysical} probe culture-specific commonsense across many languages. For Arabic specifically, ARB \citep{ghaboura2025arbcomprehensivearabicmultimodal} provides a comprehensive multimodal benchmark, and Jawaher \citep{magdy2025jawahermultidialectaldatasetarabic} offers a multidialectal collection of proverbs with explanations. In the proverb reasoning space, ePiC \citep{ghosh2022epicemployingproverbscontext} tests English proverb-to-narrative matching via BIG-bench, and ProverbEval \citep{azime2025proverbevalexploringllmevaluation} benchmarks proverb understanding in low-resource Ethiopian languages including Amharic. However, Jawaher and ProverbEval test proverb \emph{explanation}---a knowledge retrieval task---not \emph{application} to novel scenarios. ePiC targets application but is English-only. In contrast to our work, each of these benchmarks is constructed ad hoc, with no shared, replicable pipeline that communities can apply to new languages.

\paragraph{Positioning of ADAGE.}
ADAGE addresses three gaps in this landscape. First, it targets analogical \emph{application}: models must map a proverb's abstract meaning onto a novel scenario, not merely explain it; second, it provides a replicable pipeline with a principled distractor strategy, clustering proverbs by theme enables same-cluster distractors that test fine-grained reasoning, not just topic recognition; and third, we validate the pipeline across three typologically and orthographically diverse languages (Arabic, Amharic, and Japanese), demonstrating portability and flexibility. To our knowledge, ADAGE is the first methodology that enables communities of native speakers to construct rigorous, difficulty-controlled reasoning benchmarks in their own languages without translation infrastructure or ad hoc design. 

\section{Methodology}

The ADAGE pipeline constructs a culturally grounded reasoning benchmark in seven stages: (1)~data acquisition, (2)~thematic clustering, (3)~scenario generation, (4)~automated filtering, (5)~distractor selection, (6)~multi-agent validation, and (7)~dataset curation. Each stage addresses a distinct failure mode---generation targets quality, filtering targets validity, and distractor selection targets difficulty---so that the stages are decoupled and individually portable across languages. Figure~\ref{fig:pipeline} provides a visual overview. Full prompts and parameter settings are provided in the appendix; the complete pipeline is released as open-source code.\footnote{Code: \url{https://github.com/AhmedHajAhmed/adage-benchmark}}

\begin{figure}[t]
\centering
\begin{tikzpicture}[
    stage/.style={
        rectangle, rounded corners=3pt,
        draw=adageblue!80!black, fill=adageblue!8,
        minimum width=\columnwidth-0.6cm,
        minimum height=0.85cm, align=center,
        font=\scriptsize
    },
    human/.style={
        rectangle, rounded corners=3pt,
        draw=adagepeach!80!black, fill=adagepeach!10, dashed,
        minimum width=\columnwidth-0.6cm,
        minimum height=0.85cm, align=center,
        font=\scriptsize
    },
    arrow/.style={-stealth, thick, gray!45},
    slabel/.style={font=\scriptsize\bfseries, text=adageblue!45!black},
    hlabel/.style={font=\scriptsize\bfseries, text=adagepeach!45!black},
    ssub/.style={font=\tiny, text=adageblue!65!black},
    hsub/.style={font=\tiny, text=adagepeach!65!black}
]
\def\vs{1.25}
\node[stage] (s1) at (0,0) {};
\node[slabel] at (0,0.13) {1.\ Data Acquisition};
\node[ssub]   at (0,-0.15) {Proverb corpus $\to$ normalize \& deduplicate};

\node[stage] (s2) at (0,-1*\vs) {};
\node[slabel] at (0,-1*\vs+0.13) {2.\ Thematic Clustering};
\node[ssub]   at (0,-1*\vs-0.15) {Label $\to$ embed $\to$ cluster $\to$ name};

\node[stage] (s3) at (0,-2*\vs) {};
\node[slabel] at (0,-2*\vs+0.13) {3.\ Scenario Generation};
\node[ssub]   at (0,-2*\vs-0.15) {10 scenarios/proverb via GPT-OSS-120B};

\node[stage] (s4) at (0,-3*\vs) {};
\node[slabel] at (0,-3*\vs+0.13) {4.\ Automated Filtering};
\node[ssub]   at (0,-3*\vs-0.15) {Language, length, keyword, embedding checks};

\node[human] (hv) at (0,-4*\vs) {};
\node[hlabel] at (0,-4*\vs+0.13) {Human Quality Audit};
\node[hsub]   at (0,-4*\vs-0.15) {Native-speaker sanity check};

\node[stage] (s5) at (0,-5*\vs) {};
\node[slabel] at (0,-5*\vs+0.13) {5.\ Distractor Selection \& MCQ Assembly};
\node[ssub]   at (0,-5*\vs-0.15) {Same-cluster + cross-theme near-match + random};

\node[stage] (s6) at (0,-6*\vs) {};
\node[slabel] at (0,-6*\vs+0.13) {6.\ Multi-Agent Validation};
\node[ssub]   at (0,-6*\vs-0.15) {3 LLM judges $\times$ 3 rubrics};

\node[stage] (s7) at (0,-7*\vs) {};
\node[slabel] at (0,-7*\vs+0.13) {7.\ Dataset Curation};
\node[ssub]   at (0,-7*\vs-0.15) {Proverb-level train/dev/test split};

\foreach \i/\j in {s1/s2, s2/s3, s3/s4, s4/hv, hv/s5, s5/s6, s6/s7}
  \draw[arrow] (\i) -- (\j);
\end{tikzpicture}
\caption{The ADAGE pipeline. Seven stages flow top-to-bottom: (1)~acquire and deduplicate proverbs, (2)~embed and cluster by theme, (3)~generate scenarios via LLM, (4)~filter by language/length/keyword/embedding checks, (5)~assemble same-cluster, cross-theme near-match, and random distractors, (6)~validate with 3 LLM judges $\times$ 3 rubrics, (7)~split by proverb into train/dev/test. Solid boxes are automated; the dashed box (between stages 4 and 5) is a human quality audit. All stages except (1) are language-independent.}
\label{fig:pipeline}
\end{figure}
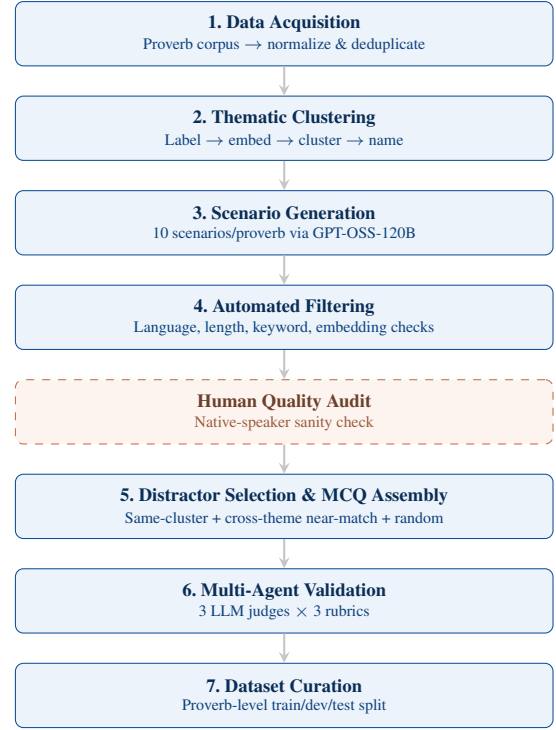

\subsection{Thematic Clustering and Distractor Strategy}
\label{sec:clustering}

The core methodological contribution of ADAGE is the use of thematic clustering to enable \emph{difficulty-by-design} distractor selection. The clustering procedure operates in three passes.

In \textbf{Pass~1}, a large language model assigns a concise English theme label (e.g., ``patience and perseverance,'' ``betrayal consequences'') to each proverb, given the proverb text and its native-language explanation. English labels are used because they serve only as intermediate clustering targets---never exposed in the final benchmark items---and the labeling model exhibits stronger instruction-following in English \citep{dey2024betteraskenglishevaluation}. In \textbf{Pass~2}, the unique theme labels are embedded using a multilingual sentence-transformer \citep{reimers-2019-sentence-bert} and grouped via agglomerative clustering with Ward's linkage. The number of clusters $k$ is selected by exhaustive search over $k \in [10, 25]$, choosing the value that maximizes the weighted silhouette coefficient \citep{ROUSSEEUW198753}.\footnote{The optimal weighted silhouette scores are low across all three languages: 0.058 (Arabic, $k{=}24$), 0.052 (Amharic, $k{=}20$), and 0.075 (Japanese, $k{=}8$). This is expected rather than problematic. Proverbs are inherently polysemous: a single proverb may encode overlapping themes (e.g., patience, humility, and social propriety), producing soft boundaries that hard cluster assignments cannot fully capture. The silhouette coefficient penalizes such overlap heavily, so low scores are a domain property rather than a pipeline failure.

Crucially, the purpose of clustering in ADAGE is not to recover ground-truth semantic categories but to generate plausible distractors. Even imperfect clusters serve this purpose: a same-cluster distractor need only be thematically \emph{related} to the correct answer, not thematically \emph{identical}. The error analysis in \S4.4 confirms that the distractor strategy functions as designed---same-cluster distractors attract substantially more errors than random distractors across all three languages---providing empirical validation that the clustering granularity is sufficient for difficulty-by-design item construction.}

In \textbf{Pass~3}, each cluster is given a descriptive name by prompting the LLM with its member labels.

Clustering on \emph{theme labels} rather than raw proverb text is a deliberate design choice. Proverb embeddings encode surface-level features alongside semantics; two proverbs expressing the same theme through different metaphors may produce distant embeddings. Theme labels abstract away this variation, producing thematically coherent groups that would be difficult to recover from text embeddings alone.

The clusters are then used for distractor selection. Each multiple-choice question (MCQ) item consists of one correct proverb and three distractors, assembled as a controlled difficulty gradient:

\begin{enumerate}
    \item \textbf{Same-cluster distractor}: a proverb from the same thematic cluster as the correct answer, creating a thematic confusion trap. Candidates with embedding similarity above 0.85 are excluded to prevent near-synonyms from appearing as distractors.
    \item \textbf{Cross-theme near-match}: the most embedding-similar proverb from the full pool (excluding same-cluster), subject to the same 0.85 cap. This traps models that rely on surface similarity as a proxy for correctness.
    \item \textbf{Random-cluster distractor}: a proverb from a different cluster, serving as a lower-plausibility control that establishes a difficulty baseline.
\end{enumerate}

\noindent Answer order is shuffled with a seeded random number generator to eliminate positional bias.

\begin{table}[t]
\centering
\small
\begin{tabular}{p{0.93\columnwidth}}
\toprule
\textbf{Scenario} \\
\midrule
\begin{CJK}{UTF8}{min}数学の先生は答案用紙を電子スキャンで二度チェックし、誤植が残っていないか慎重に校閲した上で、生徒に配布した。\end{CJK} \\[2pt]
\textit{A math teacher double-checked exam papers with an electronic scan, carefully proofreading for typos before distributing them to students.} \\
\midrule
\textbf{Which proverb best fits this scenario?} \\
\midrule
(A) \begin{CJK}{UTF8}{min}亀の甲より年の功\end{CJK} \hfill \textcolor{gray}{\scriptsize cross-theme near-match} \\
\small\textit{``Experience is worth more than a tortoise's shell.''} \\[2pt]
\textbf{(B)} \begin{CJK}{UTF8}{min}石橋を叩いて渡る\end{CJK} \hfill \textcolor{gray}{\scriptsize \textbf{correct}} \\
\small\textbf{\textit{``Tap the stone bridge before crossing.''}} \\[2pt]
(C) \begin{CJK}{UTF8}{min}一寸先は闇\end{CJK} \hfill \textcolor{gray}{\scriptsize same-cluster} \\
\small\textit{``An inch ahead is darkness.''} \\[2pt]
(D) \begin{CJK}{UTF8}{min}十人十色\end{CJK} \hfill \textcolor{gray}{\scriptsize random-cluster} \\
\small\textit{``Ten people, ten colors.''} \\
\bottomrule
\end{tabular}
\caption[Example CAPR-ja item]{Example CAPR-ja item from the \textit{Boldness vs.\ Caution} cluster. The correct proverb~(B) means ``look before you leap''; the cross-theme near-match~(A) also evokes patience and wisdom but refers to the value of experience, not caution. 4 of 7 models that answered incorrectly chose~(A).\footnotemark}
\label{tab:example-item}
\end{table}
\footnotetext{These Japanese proverbs translate naturally into familiar English equivalents---Arabic proverbs typically do not, illustrating precisely why translation-based evaluation cannot capture culturally grounded reasoning.}

\subsection{Scenario Generation and Filtering}

For each proverb, an LLM generates 10 candidate scenarios---short, modern narratives in the target language that illustrate the proverb's meaning through analogy.  The ten social domains rotated across scenarios are: workplace, family, neighbors/local community, friendship, commerce/marketplace, education, politics/public affairs, travel/diaspora, sports/competition, and technology/internet. We use GPT-OSS-120B \citep{openai2025gptoss120bgptoss20bmodel}, a 116.8B-parameter open-weight MoE model served locally via Ollama. The generation prompt enforces six constraints: concreteness, analogy rather than paraphrase, lexical separation from the proverb text, domain diversity (rotating across 10 social contexts), native-language output, and structured JSON format \footnote{The full prompt and constraint rationale are described in 
appendix Figure~\ref{fig:generation-prompt}.
}.
The prompt is written in English with a target-language output instruction, so the same template applies to all three languages without modification.

Generated scenarios are filtered automatically: each must pass a language identification check ($\geq$70\% target-script characters), a length constraint (10--120 words), a keyword overlap check ($\leq$30\% content-word overlap with the proverb), and an embedding similarity check ($\leq$0.85 cosine similarity to the proverb). Scenarios that paraphrase the proverb too closely or leak lexical cues are discarded. Between filtering and distractor selection, a native-speaker quality audit on randomly sampled scenarios (100 per language, one native speaker each) confirmed that all generated texts are coherent and culturally appropriate.
This step is a lightweight sanity check, not the primary validation; rigorous item-level quality assessment is performed by the multi-agent validation stage (\S\ref{sec:validation}).

\subsection{Cross-Linguistic Instantiation}

We validate ADAGE's portability by constructing benchmarks for three typologically diverse languages, summarized in Table~\ref{tab:benchmark-stats}.

\begin{table}[t]
\centering
\small
\resizebox{\columnwidth}{!}{%
\begin{tabular}{lrrr}
\toprule
& \textbf{Arabic} & \textbf{Amharic} & \textbf{Japanese} \\
\midrule
Seed source & Jawaher & ProverbEval & Kotowaza.org \\
Proverbs & 822 & 484 & 69 \\
Clusters & 24 & 20 & 8 \\
Scenarios generated & 8,185 & 4,809 & 690 \\
After filtering & 8,172 & 4,787 & 682 \\
Test items & 1,222 & 716 & 100 \\
\bottomrule
\end{tabular}}
\caption{Benchmark statistics for each language. All benchmarks are split 70/15/15 at the proverb level to prevent memorization of proverb--theme associations.}
\label{tab:benchmark-stats}
\end{table}

\textbf{Arabic} uses the Jawaher dataset \citep{magdy2025jawahermultidialectaldatasetarabic}, a multidialectal proverb collection with explanations in Arabic and English, yielding 822 unique proverbs after normalization and deduplication. \textbf{Amharic} draws on ProverbEval \citep{azime2025proverbevalexploringllmevaluation}, which provides 484 proverbs with meanings in Amharic and English. \textbf{Japanese} uses proverbs from Kotowaza.org, a trilingual proverb dictionary, yielding 69 entries. Because this source provides only English-language meanings (no native Japanese explanations), the theme labeling and scenario generation stages receive English explanations in the native-meaning prompt slot---a limitation that tests the pipeline's robustness when native-language metadata is unavailable. The only other language-specific adaptations are the Unicode ranges for language identification, the stopword lists for keyword overlap, and a morphological analyzer (Janome\footnote{https://janome.mocobeta.dev/}) for Japanese word segmentation. All other stages---clustering, distractor selection, validation, and evaluation---are identical across languages.

\subsection{Evaluation Protocol}

We evaluate fourteen open-weight models spanning four families and 270M--35B parameters: Gemma~3 (270M, 1B, 4B, 12B, 27B) \citep{gemmateam2025gemma3technicalreport}, Qwen~3 (1.7B, 4B, 8B, 14B, 32B) \citep{yang2025qwen3technicalreport}, Aya~23 (8B, 35B) \citep{aryabumi2024aboropenweightresearch}, and Aya~Expanse (8B, 32B) \citep{dang2024ayaexpansecombiningresearch}. All models are served locally via Ollama under zero-shot prompting: each receives a scenario and four answer options entirely in the target language, with no demonstrations or chain-of-thought instructions. GPT-OSS-120B is excluded from evaluation to avoid generator-evaluator overlap. Performance is measured by accuracy over valid responses.

For cross-benchmark comparison, we evaluate the same models on ePiC \citep{ghosh2022epicemployingproverbscontext}, an English proverb-reasoning MCQ benchmark from BIG-bench that tests the same cognitive task---matching narratives to proverbs---in an English cultural context.

\subsection{Multi-Agent Validation}
\label{sec:validation}

Benchmark quality is assessed by a panel of three open-weight LLM judges---Gemma~3 27B, Mistral~Small~3.2, and Aya~Expanse~32B---selected for provider diversity and multilingual coverage. Each judge scores every item on three rubrics (scenario faithfulness, naturalness, distractor plausibility) using a 1--5 Likert scale.
The rubric definitions and full prompts are in Appendix (Figure~\ref{fig:validation-prompt}). 
To screen for quality, items are flagged if their mean score falls below 3.0 on any rubric or if inter-judge disagreement exceeds 2 points. 

\section{Results}

\subsection{Benchmark Quality}

Table~\ref{tab:quality} reports majority-vote pass rates (2 of 3 judges rating ``acceptable'') across all three benchmarks. Arabic and Japanese achieve high pass rates on all rubrics ($\geq$98.7\%), with flagging rates of 4.4\% and 0.7\% respectively. Amharic shows a different pattern: faithfulness and distractor plausibility pass at 100\%, but naturalness drops to 77.7\%, and the overall flagging rate is 35.3\%. This is driven by two factors: 367 items (7.7\%) where all three judges failed to produce valid evaluations---likely reflecting weaker Amharic-language processing by the judge models---and high inter-judge disagreement on naturalness, where Gemma~3 rates substantially lower than Mistral and Aya (mean 3.28 vs.\ 4.92 and 3.59). This is consistent with Amharic's status as a genuinely low-resource language for current LLMs, affecting judge reliability and generation quality. Flagged items are retained in the final benchmarks to preserve test set size, particularly for Amharic (37.6\% of test items are flagged) and to avoid selection bias. As a robustness check, we also report accuracy on the unflagged subset in \S\ref{sec:robustness}; results are consistent with the full-set evaluation.

\begin{table}[t]
\centering
\small
\begin{tabular}{lccc}
\toprule
\textbf{Rubric} & \textbf{ar} & \textbf{am} & \textbf{ja} \\
\midrule
Faithfulness ($\geq$4) & 98.7 & 100.0 & 98.7 \\
Naturalness ($\geq$4)  & 100.0 & 77.7 & 100.0 \\
Distractors ($\geq$3)  & 99.4 & 100.0 & 99.9 \\
\midrule
Flagged (\%)           & 4.4 & 35.3 & 0.7 \\
\bottomrule
\end{tabular}
\caption{Majority-vote pass rates (\%) and flagging rates per language. Rubric column parenthetical values are the minimum Likert score (1--5) for a judge to rate an item acceptable. The Amharic naturalness gap reflects weaker judge reliability on low-resource languages.}
\label{tab:quality}
\end{table}

\noindent Inter-annotator agreement (IAA) on raw scores is low across all languages (Krippendorff's $\alpha$ ranges from $-0.42$ to $0.32$), but this reflects calibration differences---judges use the 1--5 scale at different offsets---rather than genuine disagreement about which items are good or bad. When each judge's scores are standardized to zero mean and unit variance ($z$-score normalization), all negative $\alpha$ values disappear. Similarly, binarized agreement (Cohen's $\kappa$) appears low not because judges disagree, but because nearly all items pass---when one category dominates, $\kappa$ is deflated even at high raw agreement rates (the prevalence paradox). 
\footnote{Appendix Tables~\ref{tab:iaa-primary}--\ref{tab:iaa-spearman} report the full IAA results across all three languages.}

\subsection{Overall Accuracy}

Table~\ref{tab:cross-benchmark} reports accuracy for all 14 models on ePiC (English) and the three CAPR benchmarks. Several patterns emerge.

\begin{table*}[t]
\centering
\small
\begin{tabular}{llrrrr}
\toprule
\textbf{Model} & \textbf{Params} & \textbf{ePiC-en (\%)} & \textbf{CAPR-ar (\%)} & \textbf{CAPR-am (\%)} & \textbf{CAPR-ja (\%)} \\
\midrule
Gemma 3 27B       & 27B  & 83.0 & 70.1 & 41.3 & 86.0 \\
Qwen 3 32B        & 32B  & 82.0 & 67.4 & 30.7 & 89.0 \\
Qwen 3 14B        & 14B  & 82.5 & 66.4 & 32.4 & 91.0 \\
Qwen 3 8B         & 8B   & 79.8 & 65.5 & 28.9 & 77.0 \\
Gemma 3 12B       & 12B  & 81.3 & 65.3 & 39.2 & 85.0 \\
Aya Expanse 32B   & 32B  & 77.5 & 63.3 & 32.7 & 93.0 \\
Qwen 3 4B         & 4B   & 77.8 & 58.1 & 25.8 & 77.0 \\
Aya 35B           & 35B  & 74.7 & 54.5 & 25.0 & 80.0 \\
Aya Expanse 8B    & 8B   & 69.0 & 53.5 & 25.4 & 76.0 \\
Gemma 3 4B        & 4B   & 62.4 & 50.6 & 30.3 & 55.0 \\
Qwen 3 1.7B       & 1.7B & 57.4 & 44.7 & 29.6 & 38.0 \\
Aya 8B            & 8B   & 58.5 & 44.4 & 25.1 & 62.0 \\
Gemma 3 1B        & 1B   & 32.0 & 30.0 & 27.8 & 28.0 \\
Gemma 3 270M      & 270M & 18.4 & 24.5 & 24.0 & 22.0 \\
\midrule
Random baseline   & ---  & 20.0 & 25.0 & 25.0 & 25.0 \\
\bottomrule
\end{tabular}
\caption{Accuracy on ePiC (English, 5-choice) and the three CAPR benchmarks (4-choice), sorted by CAPR-ar accuracy. All evaluations are zero-shot. The random baseline differs because ePiC uses 5 answer choices.}
\label{tab:cross-benchmark}
\end{table*}

First, all three CAPR benchmarks are non-trivial: the best model achieves 70.1\% on Arabic, 41.3\% on Amharic, and 93.0\% on Japanese. Amharic accuracy barely exceeds chance (25\%), confirming it as a genuinely hard low-resource benchmark; Japanese accuracy is notably higher (see Discussion). The smallest model (Gemma~3 270M) performs at or near chance on all benchmarks, confirming that the tasks cannot be solved by positional heuristics. Second, no single model family dominates: Gemma~3 leads on Arabic, but Qwen~3 is competitive at multiple scales, and Aya~Expanse consistently outperforms the older Aya models at matched sizes (e.g., 63.3\% vs.\ 54.5\% at 32--35B on Arabic), indicating that targeted multilingual training improvements contribute meaningfully to culturally grounded reasoning.

Third, the cultural reasoning gap is consistent. Models that perform well on English proverb reasoning drop substantially on native benchmarks. On Arabic, accuracy falls by 10--20 percentage points relative to ePiC for the strongest models; on Amharic, the drop is 32--52 pp, with most models near chance. Japanese shows smaller gaps, discussed below.

\subsection{Scaling Behavior}

\begin{figure*}[t]
\centering
\includegraphics[width=\textwidth]{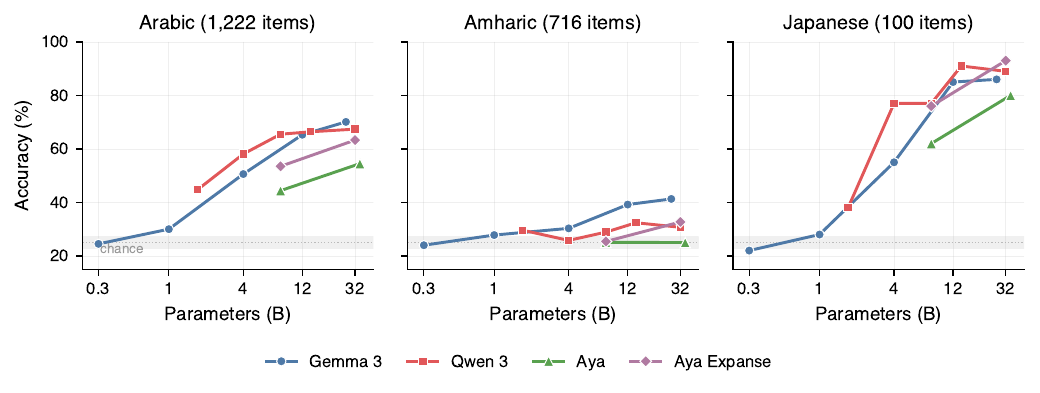}
\caption{Accuracy vs.\ model size (log scale) across three languages. Dashed line = 25\% chance. Scaling yields clear gains on Arabic and Japanese but negligible improvement on Amharic, where most models cluster near chance regardless of size.}
\label{fig:scaling}
\end{figure*}

Figure~\ref{fig:scaling} plots accuracy against model size (log scale) for all four families across the three languages.

Within each family, accuracy increases with model size (Figure~\ref{fig:scaling}), but with diminishing returns and family-specific ceilings. Gemma~3 shows the steepest scaling, rising from chance at 270M to 70.1\% at 27B on Arabic---approximately log-linear across five sizes. Qwen~3 scales sharply from 1.7B to 8B (+20.8 pp on Arabic) but then plateaus: 8B to 32B yields only 1.9 pp. This pattern suggests that scaling within a fixed architecture and training distribution encounters diminishing returns on culturally grounded reasoning well before the parameter frontier.

The Aya family reveals that training data composition matters as much as scale. Aya~Expanse 8B outperforms the original Aya 8B by 9.1 pp on Arabic (53.5\% vs.\ 44.4\%), a gain comparable to quadrupling parameters within the Gemma family. On Amharic, scaling effects largely vanish: all models cluster near chance (24--41\%), suggesting that Amharic-language competence is the binding constraint rather than reasoning capacity.

\subsection{Error Analysis}

\paragraph{Distractor type vulnerability.} Table~\ref{tab:distractor-errors} reports the distribution of errors by distractor type. On Arabic, cross-theme near-matches attract the largest share of errors (34.4\%), followed by same-cluster (31.2\%) and random-cluster (26.9\%)---mirroring the intended difficulty gradient. This gradient sharpens with model capability: Gemma~3 27B shows a 16.2~pp gap between cross-theme near-match and random-cluster error rates, while Gemma~3 270M shows a flat distribution consistent with random guessing. On Japanese, a similar pattern holds with random-cluster distractors attracting the fewest errors (30.1\%). On Amharic, however, the distribution is nearly flat (32.9\%/34.7\%/32.4\%), consistent with models performing near chance---the difficulty gradient cannot emerge when models lack the linguistic competence to engage with the task content. 
\footnote{Appendix Tables~\ref{tab:cluster-ar}, \ref{tab:cluster-am}, and~\ref{tab:cluster-ja} report accuracy averaged across all 14 models for each thematic cluster on CAPR-ar, CAPR-am, and CAPR-ja respectively.}

\begin{table}[t]
\centering
\small
\begin{tabular}{lrrr}
\toprule
\textbf{Distractor type} & \textbf{ar (\%)} & \textbf{am (\%)} & \textbf{ja (\%)} \\
\midrule
Cross-theme near-match    & 34.4 & 32.9 & 34.2 \\
Same cluster     & 31.2 & 34.7 & 35.8 \\
Random cluster   & 26.9 & 32.4 & 30.1 \\
\bottomrule
\end{tabular}
\caption{Aggregate error distribution by distractor type. The gradient mirrors the intended difficulty ordering on Arabic and Japanese; Amharic is flat, consistent with near-chance performance.}
\label{tab:distractor-errors}
\end{table}

\paragraph{Error convergence.} When multiple models answer an item incorrectly, they tend to converge on the \emph{same} wrong distractor. We measure this as the proportion of wrong answers that fall on the most-chosen distractor for each item, averaged across items (1.0 = all wrong models pick the same option; 0.33 = uniform spread). The mean convergence score is 0.652 on Arabic, 0.616 on Japanese, and 0.527 on Amharic---all substantially above the 0.33 uniform baseline. Even on Amharic, where accuracy is near chance, convergence exceeds 0.50, indicating that distractor design systematically exposes shared weaknesses in how LLMs represent proverb semantics rather than eliciting random noise. 

\paragraph{Position bias.} Position bias severity correlates inversely with model size. Models above 4B parameters show near-uniform answer distributions, while Gemma~3 270M selects option D in 89.6\% of cases, confirming that it has not learned the MCQ format and defaults to positional heuristics. 
\footnote{Exhaustive position bias data are in appendix Tables~\ref{tab:pos-ar}, \ref{tab:pos-am}, and \ref{tab:pos-ja}}. 

\paragraph{Robustness to item quality.}
\label{sec:robustness}
To verify that flagged items do not distort the main findings, we re-evaluate the top 5 models on the unflagged subsets of each test set (Arabic: 1,166/1,222 items; Amharic: 447/716; Japanese: 100/100---zero items flagged).
Accuracy on the unflagged subsets is within 1.3 pp of the full-set results across all models and languages, confirming that the cultural reasoning gap is not an artifact of item quality variation. The largest shift is Qwen~3 32B on Amharic (+1.3 pp), despite removing 37.6\% of test items.

\section{Discussion}

The most striking result is the accuracy spread across languages: 41.3\% (Amharic), 70.1\% (Arabic), 93.0\% (Japanese). This variation is not a pipeline artifact; the same seven stages, the same models, and the same evaluation protocol produce dramatically different outcomes depending on the target language. This variation validates the language-agnostic framing: because each benchmark tests culturally native content rather than translated English reasoning, the difficulty differences reflect genuine gaps in language-specific competence that translation-based evaluation would obscure. The three-language comparison also reveals that ``multilingual'' model performance is not a single capability but a composite of language-specific competences that vary by orders of magnitude within the same model.

Amharic is the most revealing case. With the best model at 41.3\% (chance = 25\%), a flat distractor gradient (32.9\%/34.7\%/32.4\%), extreme position bias even in large models ($\chi^2 > 100$ for Gemma~3 27B and Qwen~3 32B), and near-zero cross-model agreement ($\rho = 0.048$), the evidence points to a \textit{language competence} failure rather than a reasoning failure. Models cannot engage with Amharic text reliably enough for the designed difficulty gradient to emerge. This is itself a meaningful finding: current ``multilingual'' models have essentially no culturally grounded reasoning capability in genuinely low-resource languages, a gap that translation-based evaluation would never reveal.

Japanese accuracy is notably high, but three factors temper its interpretation. First, the test set contains only 100 items across 7 clusters, reducing statistical power and thematic diversity relative to Arabic (1,222 items, 23 clusters). Second, Japanese is a high-resource language with extensive pretraining coverage, so strong performance may reflect richer proverb representations in training data. Third, some Japanese proverbs---for example, \begin{CJK}{UTF8}{min}一石二鳥\end{CJK} (``one stone, two birds'')---have well-known Chinese or English cognates, enabling cross-lingual transfer that is unavailable for Arabic or Amharic proverbs. These factors do not invalidate the Japanese benchmark but suggest it sets a lower ceiling for discriminative power; expanding the test set is a clear next step.

The distractor strategy and scaling behavior interact with language competence in a consistent pattern. On Arabic, where models engage meaningfully with the content, the intended difficulty gradient emerges clearly and sharpens with scale. On Japanese, where accuracy is high, the gradient is attenuated but random-cluster distractors remain the least deceptive. On Amharic, the gradient vanishes entirely. Similarly, scaling yields log-linear gains on Arabic, diminishing returns on Japanese, and negligible improvement on Amharic (Figure~\ref{fig:scaling})---suggesting that parameter count helps only after a minimum threshold of language competence is met. Training data composition matters at least as much: Aya~Expanse's targeted multilingual training yields gains comparable to quadrupling parameters within the Gemma family on Arabic, yet provides no advantage on Amharic.

\section{Conclusion}

We introduced ADAGE, a language-agnostic pipeline for constructing culturally grounded reasoning benchmarks without translation. By combining native-speaker curation with LLM-assisted generation and a cluster-based distractor strategy, ADAGE produces difficulty-controlled MCQ benchmarks that test abstract analogical reasoning as it is expressed natively in each language. We validated the pipeline's portability by constructing benchmarks for three typologically diverse languages---Arabic, Amharic, and Japanese---and evaluating 14 open-weight models under zero-shot conditions.

The results reveal a cultural reasoning gap that varies dramatically with language resource level. On Arabic, the best model reaches 70.1\%---substantially below human performance---with a clear distractor difficulty gradient confirming that failures are systematic. On Amharic, accuracy barely exceeds chance (41.3\%), with flat distractor distributions and extreme position bias indicating that current models lack the language competence to engage with the task. On Japanese, accuracy reaches 93.0\%, consistent with high pretraining coverage but constrained by a smaller test set. Across all three languages, models that achieve strong performance on English proverb reasoning drop by 12--52 percentage points on native benchmarks, and this gap does not close with scale alone.

These findings demonstrate that evaluation-by-translation substantially overestimates multilingual reasoning capabilities. We release the pipeline, benchmarks, and evaluation suite to support expansion to additional languages.

\section*{Limitations}

\paragraph{Test set size imbalance.} The three benchmarks differ substantially in size: 1,222 items (Arabic), 716 (Amharic), and 100 (Japanese). The Japanese test set in particular limits statistical power and cluster coverage (7 of 8 clusters represented). Cross-language comparisons should be interpreted with this asymmetry in mind.

\paragraph{Zero-shot evaluation only.} All models were evaluated under zero-shot prompting with a 300-second timeout. Qwen~3 models, which use chain-of-thought reasoning internally, produced a small number of unparseable responses---up to 3.8\% on Amharic---likely due to exceeding this limit (details are described in Appendix~\ref{sec:compute}). Few-shot or chain-of-thought strategies may unlock additional capability, and the gap between prompted and unprompted performance would itself be diagnostic---large improvements would suggest latent cultural knowledge that models fail to access spontaneously.

\paragraph{Open-weight models only.} The evaluation covers models up to 35B parameters. Larger proprietary models (GPT, Claude, Gemini) may exhibit qualitatively different performance, and their inclusion would provide a more complete picture of the current frontier.

\paragraph{LLM-generated scenarios and LLM judges.} Both scenario generation and multi-agent validation rely on LLMs. While the human quality audit and high majority-vote pass rates (98.7--100\% on Arabic and Japanese) provide evidence of quality, systematic human annotation would strengthen confidence, particularly for Amharic where judge reliability is lower (35.3\% flagging rate).

\paragraph{Cross-benchmark comparison.} The English baseline (ePiC) differs from CAPR in format (5 vs.\ 4 options), scenario source (crowdsourced vs.\ LLM-generated), and human baseline (70\% from unscreened MTurk workers). These differences arguably strengthen the comparison---the gap cannot be attributed to shared construction artifacts---but direct comparability is limited.

\section*{Ethical Considerations}

The benchmarks contain proverbs that reference violence, gender roles, religion, and social hierarchy. These reflect authentic cultural wisdom traditions and are retained without sanitization to preserve the reasoning challenges they encode. No scenarios depict graphic content; all are set in everyday modern contexts. English translations of some proverbs may appear jarring when removed from their cultural context.


\bibliography{custom}
\newpage
\clearpage
\appendix

\section{Computational Resources and Prompts}
\subsection{Compute Environment}
\label{sec:compute}

All generation, validation, and evaluation were performed locally on a Mac Studio with an Apple M3~Ultra (32-core CPU, 80-core GPU, 512\,GB unified memory). Models were served via Ollama using each model's native chat template. GPT-OSS-120B (116.8B parameters, MoE with 5.1B active parameters per forward pass) was loaded in its native MXFP4 quantization (60.8\,GiB) without CPU offloading. Generation used temperature 1.0; validation judges used temperature 0.3; evaluation used temperature 0.0. No proprietary APIs were used at any pipeline stage.

\subsection{Replication Details}
\begin{figure*}[h!]
\begin{tcolorbox}[colback=gray!5, colframe=gray!60, title={\textbf{System Prompt}}, fonttitle=\small, fontupper=\small\ttfamily, left=4pt, right=4pt, top=4pt, bottom=4pt]
You are an expert evaluator of educational assessment items. You will evaluate a multiple-choice question (MCQ) that tests understanding of proverbs through real-world scenarios.\\[4pt]
You must evaluate the item on exactly three rubrics, each scored from 1 (worst) to 5 (best):\\[4pt]
1. \textbf{Scenario Faithfulness}:\ignorespaces\\
\hspace*{1em}5: The scenario perfectly illustrates the proverb's meaning through analogy. The connection is clear but requires inference --- it is NOT a paraphrase.\\
\hspace*{1em}4: Illustrates meaning well, with minor ambiguity.\\
\hspace*{1em}3: Partially captures meaning but analogy is weak or unclear.\\
\hspace*{1em}2: Only loosely relates; connection is forced or misleading.\\
\hspace*{1em}1: No meaningful connection, or directly paraphrases the proverb.\\[4pt]
2. \textbf{Scenario Naturalness}:\\
\hspace*{1em}5: Natural, coherent, well-written narrative that reads like a real-world situation.\\
\hspace*{1em}4: Mostly natural with minor awkwardness.\\
\hspace*{1em}3: Understandable but noticeably unnatural phrasing.\\
\hspace*{1em}2: Significant language issues (grammar, coherence, register).\\
\hspace*{1em}1: Largely incoherent, ungrammatical, or unreadable.\\[4pt]
3. \textbf{Distractor Plausibility}:\\
\hspace*{1em}5: All 3 wrong answers are plausible but distinguishable by a knowledgeable reader.\\
\hspace*{1em}4: 2--3 distractors plausible; at most one obviously wrong.\\
\hspace*{1em}3: At least 1 distractor plausible; others somewhat easy to eliminate.\\
\hspace*{1em}2: Distractors mostly implausible; correct answer obvious.\\
\hspace*{1em}1: Distractors completely unrelated; item is trivial.\\[4pt]
Respond in the exact JSON format specified. Be rigorous and consistent in your scoring.
\end{tcolorbox}
\vspace{-4pt}
\begin{tcolorbox}[colback=gray!5, colframe=gray!60, title={\textbf{User Prompt}}, fonttitle=\small, fontupper=\small\ttfamily, left=4pt, right=4pt, top=4pt, bottom=4pt]
Evaluate the following proverb MCQ item.\\[4pt]
\textbf{Scenario:} \{scenario\}\\[4pt]
\textbf{Options:}\\
(A) \{option\_a\}\\
(B) \{option\_b\}\\
(C) \{option\_c\}\\
(D) \{option\_d\}\\[4pt]
\textbf{Correct Answer:} (\{correct\_letter\}) \{correct\_text\}\\[4pt]
Provide your evaluation:
\end{tcolorbox}
\caption{Validation prompt with rubric definitions (verbatim). Structured JSON output is enforced via Ollama's schema-constrained decoding.}
\label{fig:validation-prompt}
\end{figure*}


\paragraph{Invalid responses.} All models were evaluated with a 300-second generation timeout. Qwen~3 models produced a small number of unparseable responses (no valid A/B/C/D extracted), likely due to extended chain-of-thought reasoning exceeding this limit. No other model family exhibited this behavior. Affected counts: Qwen~3~8B on Amharic (27/716, 3.8\%), Qwen~3~4B on Amharic (23/716, 3.2\%), Qwen~3~1.7B on Amharic (18/716, 2.5\%), Qwen~3~14B on Amharic (7/716, 1.0\%), Qwen~3~32B on Arabic (1/1,222), Amharic (2/716), and Japanese (1/100), and Qwen~3~4B on Japanese (1/100). Invalid responses were scored as incorrect.


\begin{figure*}[h!]
\begin{tcolorbox}[colback=gray!5, colframe=gray!60, title={\textbf{System Prompt}}, fonttitle=\small, fontupper=\small\ttfamily, left=4pt, right=4pt, top=4pt, bottom=4pt]
You are an expert in \{culture\} culture and proverbs. Given a proverb with its meaning, generate \{n\} short real-world scenarios where this proverb applies.\\[4pt]
Rules:\\
1. Each scenario must be a concrete, specific story --- not an abstract description.\\
2. Each scenario must express the proverb's meaning through ANALOGY, not paraphrase. Do not restate or explain the proverb's meaning directly. The reader should need to infer the connection.\\
3. Do NOT reuse any content words (nouns, verbs, adjectives) from the proverb text.\\
4. Each scenario must come from a DIFFERENT social context. Use these in order: \{domains\}\\
5. Write in \{language\}. Each scenario: \{min\_words\}--\{max\_words\} words.\\
6. Return ONLY a JSON array of strings. No other text.
\end{tcolorbox}
\vspace{-4pt}
\begin{tcolorbox}[colback=gray!5, colframe=gray!60, title={\textbf{User Prompt}}, fonttitle=\small, fontupper=\small\ttfamily, left=4pt, right=4pt, top=4pt, bottom=4pt]
Proverb: \{proverb\}\\
Meaning (\{language\}): \{native\_explanation\}\\
Meaning (English): \{en\_explanation\}\\
English equivalent: \{en\_equivalent\}\\[4pt]
Generate \{n\} scenarios:
\end{tcolorbox}
\caption{Scenario generation prompt (verbatim). The same template is used for all three languages; only the culture descriptor, language name, and word-count bounds change.}
\label{fig:generation-prompt}
\end{figure*}

\clearpage
\newpage
\section{Supplemental Results}
\begin{table}[!htbp]
\centering
\small
\begin{tabular}{lrrrrl}
\toprule
\textbf{Model} & \textbf{A} & \textbf{B} & \textbf{C} & \textbf{D} & \textbf{$\chi^2$} \\
\midrule
Qwen 3 14B          & 145 & 210 & 183 & 171 & 12.3 \\
Qwen 3 8B           & 130 & 196 & 181 & 182 & 14.6 \\
Gemma 3 12B         & 182 & 129 & 231 & 174 & 29.3 \\
Qwen 3 1.7B         & 168 & 179 & 238 & 113 & 45.1 \\
Qwen 3 4B           & 131 & 116 & 221 & 225 & 57.8 \\
Qwen 3 32B          &  95 & 124 & 222 & 273 & 116.3 \\
Gemma 3 27B         &  75 & 229 & 266 & 146 & 122.8 \\
Gemma 3 4B          & 134 & 330 & 190 &  62 & 215.8 \\
Aya Expanse 32B     &  45 & 345 & 233 &  93 & 311.9 \\
Gemma 3 1B          & 316 &  63 & 337 &   0 & 498.5 \\
Aya 35B             & 136 & 132 & 429 &  22 & 507.4 \\
Aya Expanse 8B      &   7 & 480 & 226 &   3 & 856.8 \\
Gemma 3 270M        &   0 &  40 &  72 & 604 & 1,360.0 \\
Aya 8B              &  43 &  62 & 611 &   0 & 1,401.4 \\
\bottomrule
\end{tabular}
\caption{Position bias on CAPR-am (716 items). Ground truth: A=173, B=185, C=187, D=171.}
\label{tab:pos-am}
\end{table}

\begin{table}[!htbp]
\centering
\small
\begin{tabular}{lrrrrl}
\toprule
\textbf{Model} & \textbf{A} & \textbf{B} & \textbf{C} & \textbf{D} & \textbf{$\chi^2$} \\
\midrule
Qwen 3 4B           & 23 & 23 & 26 & 27 & 0.5 \\
Aya Expanse 32B     & 26 & 24 & 28 & 22 & 0.8 \\
Qwen 3 14B          & 24 & 24 & 29 & 23 & 0.9 \\
Aya Expanse 8B      & 23 & 25 & 29 & 23 & 1.0 \\
Qwen 3 32B          & 28 & 24 & 28 & 19 & 2.2 \\
Gemma 3 12B         & 32 & 26 & 22 & 20 & 3.4 \\
Qwen 3 8B           & 27 & 27 & 29 & 17 & 3.5 \\
Gemma 3 27B         & 33 & 25 & 24 & 18 & 4.6 \\
Gemma 3 4B          & 27 & 32 & 26 & 15 & 6.2 \\
Qwen 3 1.7B         & 21 & 30 & 34 & 15 & 8.9 \\
Aya 35B             & 19 & 24 & 38 & 19 & 9.7 \\
Aya 8B              & 38 & 20 & 27 & 15 & 11.9 \\
Gemma 3 1B          & 45 & 18 & 35 &  2 & 43.1 \\
Gemma 3 270M        &  0 &  2 &  0 & 98 & 284.3 \\
\bottomrule
\end{tabular}
\caption{Position bias on CAPR-ja (100 items). Ground truth: A=26, B=25, C=27, D=22.}
\label{tab:pos-ja}
\end{table}

\begin{table}[H]
\centering
\small
\begin{tabular}{llccc}
\toprule
\textbf{Rubric} & \textbf{Lang} & \textbf{Kripp.\ $\alpha$} & \textbf{Fleiss $\kappa$} & \textbf{ICC}\\ 
\midrule
\multirow{3}{*}{Faithfulness}
  & ar & 0.25 & 0.19 & 0.28 \\
  & am & $-$0.22 & $-$0.10 & 0.02 \\
  & ja & 0.32 & 0.24 & 0.34 \\
\midrule
\multirow{3}{*}{Naturalness}
  & ar & $-$0.24 & $-$0.22 & 0.03 \\
  & am & $-$0.33 & $-$0.20 & 0.01 \\
  & ja & $-$0.19 & $-$0.19 & 0.04 \\
\midrule
\multirow{3}{*}{Distractors}
  & ar & $-$0.14 & $-$0.14 & 0.07 \\
  & am & $-$0.42 & $-$0.24 & 0.00 \\
  & ja & $-$0.22 & $-$0.22 & 0.04 \\
\bottomrule
\end{tabular}
\caption{Primary inter-annotator agreement metrics per rubric and language.}
\label{tab:iaa-primary}
\end{table}

\begin{table}[!htbp]
\centering
\small
\begin{tabular}{llccc}
\toprule
\textbf{Rubric} & \textbf{Lang} & \textbf{Gemma 3} & \textbf{Mistral Sm.} & \textbf{Aya Exp.} \\
\midrule
\multirow{3}{*}{Faith.}
  & ar & 4.09 \tiny{(.40)} & 4.23 \tiny{(.53)} & 4.42 \tiny{(.52)} \\
  & am & 3.37 \tiny{(.61)} & 4.52 \tiny{(.50)} & 4.51 \tiny{(.50)} \\
  & ja & 4.13 \tiny{(.41)} & 4.44 \tiny{(.63)} & 4.42 \tiny{(.52)} \\
\midrule
\multirow{3}{*}{Natur.}
  & ar & 4.52 \tiny{(.50)} & 4.98 \tiny{(.15)} & 4.00 \tiny{(.39)} \\
  & am & 3.28 \tiny{(.79)} & 4.92 \tiny{(.28)} & 3.59 \tiny{(.50)} \\
  & ja & 4.82 \tiny{(.38)} & 4.98 \tiny{(.13)} & 4.22 \tiny{(.45)} \\
\midrule
\multirow{3}{*}{Distr.}
  & ar & 2.88 \tiny{(.40)} & 3.75 \tiny{(.46)} & 3.71 \tiny{(.58)} \\
  & am & 2.38 \tiny{(.48)} & 3.98 \tiny{(.22)} & 4.23 \tiny{(.44)} \\
  & ja & 3.04 \tiny{(.25)} & 3.91 \tiny{(.33)} & 3.75 \tiny{(.52)} \\
\bottomrule
\end{tabular}
\caption{Per-judge scoring means (std) by rubric and language.}
\label{tab:iaa-perjudge}
\end{table}

\begin{table}[!htbp]
\centering
\small
\begin{tabular}{llccc}
\toprule
\textbf{Rubric} & \textbf{Lang} & \textbf{Raw $\alpha$} & \textbf{$z$-Norm $\alpha$} & \textbf{$\Delta$} \\
\midrule
\multirow{3}{*}{Faith.}
  & ar & 0.25 & 0.32 & +0.07 \\
  & am & $-$0.22 & 0.05 & +0.27 \\
  & ja & 0.32 & 0.39 & +0.08 \\
\midrule
\multirow{3}{*}{Natur.}
  & ar & $-$0.24 & 0.09 & +0.33 \\
  & am & $-$0.33 & 0.06 & +0.39 \\
  & ja & $-$0.19 & 0.08 & +0.26 \\
\midrule
\multirow{3}{*}{Distr.}
  & ar & $-$0.14 & 0.15 & +0.29 \\
  & am & $-$0.42 & 0.00 & +0.42 \\
  & ja & $-$0.22 & 0.11 & +0.33 \\
\bottomrule
\end{tabular}
\caption{Effect of $z$-score normalization on Krippendorff's $\alpha$. Normalization eliminates all negative values across all languages.}
\label{tab:iaa-znorm}
\end{table}

\begin{table}[!htbp]
\centering
\small
\begin{tabular}{llccc}
\toprule
\textbf{Rubric} & \textbf{Lang} & \textbf{Gem.--Mis.} & \textbf{Gem.--Aya} & \textbf{Mis.--Aya} \\
\midrule
\multirow{3}{*}{Faith.}
  & ar & 0.31 & 0.34 & 0.30 \\
  & am & 0.04 & 0.06 & 0.03 \\
  & ja & 0.40 & 0.39 & 0.41 \\
\midrule
\multirow{3}{*}{Natur.}
  & ar & 0.07 & 0.15 & 0.07 \\
  & am & 0.08 & 0.04 & 0.04 \\
  & ja & 0.05 & 0.20 & $-$0.04 \\
\midrule
\multirow{3}{*}{Distr.}
  & ar & 0.16 & 0.17 & 0.10 \\
  & am & 0.01 & 0.00 & 0.00 \\
  & ja & 0.14 & 0.10 & 0.10 \\
\bottomrule
\end{tabular}
\caption{Pairwise Spearman $\rho$ per rubric. Arabic and Japanese faithfulness correlations are moderate ($\rho \approx 0.3$--$0.4$, $p < 0.001$); Amharic correlations are near zero, reflecting weaker judge reliability on low-resource text.}
\label{tab:iaa-spearman}
\end{table}

\begin{table}[h!]
\centering
\small
\begin{tabular}{lrrrrl}
\toprule
\textbf{Model} & \textbf{A} & \textbf{B} & \textbf{C} & \textbf{D} & \textbf{$\chi^2$} \\
\midrule
Qwen 3 8B           & 283 & 312 & 304 & 323 & 2.8 \\
Gemma 3 12B         & 330 & 267 & 317 & 308 & 7.3 \\
Qwen 3 4B           & 305 & 278 & 348 & 291 & 9.1 \\
Gemma 3 27B         & 258 & 335 & 327 & 302 & 11.8 \\
Aya Expanse 8B      & 255 & 330 & 330 & 307 & 12.3 \\
Qwen 3 14B          & 275 & 289 & 287 & 371 & 19.1 \\
Aya 35B             & 374 & 261 & 263 & 324 & 28.9 \\
Qwen 3 1.7B         & 240 & 324 & 397 & 261 & 49.1 \\
Qwen 3 32B          & 236 & 280 & 298 & 407 & 51.9 \\
Aya Expanse 32B     & 375 & 353 & 282 & 212 & 53.6 \\
Gemma 3 4B          & 311 & 414 & 351 & 146 & 128.7 \\
Aya 8B              & 410 & 445 & 202 & 165 & 199.1 \\
Gemma 3 1B          & 526 & 148 & 517 &  31 & 633.4 \\
Gemma 3 270M        &   3 &  89 &  35 & 1,095 & 2,732.8 \\
\bottomrule
\end{tabular}
\caption{Position bias on CAPR-ar (1,222 items). Ground truth: A=321, B=308, C=287, D=306.}
\label{tab:pos-ar}
\end{table}

\begin{table}[!htbp]
\centering
\small
\begin{tabular}{llr}
\toprule
\textbf{Model 1} & \textbf{Model 2} & \textbf{$\rho$} \\
\midrule
\multicolumn{3}{l}{\textit{Highest agreement}} \\
Qwen 3 14B      & Qwen 3 32B      & 0.549 \\
Gemma 3 27B     & Qwen 3 14B      & 0.517 \\
Gemma 3 12B     & Gemma 3 27B     & 0.512 \\
Qwen 3 14B      & Qwen 3 8B       & 0.502 \\
Gemma 3 27B     & Qwen 3 32B      & 0.500 \\
\midrule
\multicolumn{3}{l}{\textit{Lowest agreement}} \\
Gemma 3 270M    & Qwen 3 4B       & $-$0.065 \\
Aya 8B          & Gemma 3 270M    & $-$0.113 \\
Aya Exp.\ 32B   & Gemma 3 270M    & $-$0.151 \\
Gemma 3 270M    & Gemma 3 4B      & $-$0.218 \\
Gemma 3 1B      & Gemma 3 270M    & $-$0.331 \\
\bottomrule
\end{tabular}
\caption{Top-5 and bottom-5 pairwise Spearman $\rho$ of item-level correctness on CAPR-ar. Mean $\rho$ across all 91 pairs: 0.257.}
\label{tab:cross-agree-ar}
\end{table}

\begin{table}[!htbp]
\centering
\small
\begin{tabular}{llr}
\toprule
\textbf{Model 1} & \textbf{Model 2} & \textbf{$\rho$} \\
\midrule
\multicolumn{3}{l}{\textit{Highest agreement}} \\
Gemma 3 27B & Gemma 3 12B & 0.313 \\
Aya Exp.\ 32B & Aya Exp.\ 8B & 0.291 \\
Aya 8B & Gemma 3 1B & 0.244 \\
Gemma 3 27B & Gemma 3 4B & 0.199 \\
Gemma 3 12B & Gemma 3 4B & 0.161 \\
\midrule
\multicolumn{3}{l}{\textit{Lowest agreement}} \\
Qwen 3 1.7B & Gemma 3 270M & $-$0.107 \\
Gemma 3 4B & Gemma 3 270M & $-$0.165 \\
Aya 8B & Gemma 3 270M & $-$0.198 \\
Aya Exp.\ 8B & Gemma 3 270M & $-$0.201 \\
Gemma 3 1B & Gemma 3 270M & $-$0.247 \\
\bottomrule
\end{tabular}
\caption{Top-5 and bottom-5 pairwise Spearman $\rho$ of item-level correctness on CAPR-am. Mean $\rho$: 0.048.}
\label{tab:cross-agree-am}
\end{table}

\begin{table}[!htbp]
\centering
\small
\begin{tabular}{llr}
\toprule
\textbf{Model 1} & \textbf{Model 2} & \textbf{$\rho$} \\
\midrule
\multicolumn{3}{l}{\textit{Highest agreement}} \\
Qwen 3 32B & Gemma 3 12B & 0.568 \\
Gemma 3 27B & Gemma 3 12B & 0.557 \\
Qwen 3 32B & Aya Exp.\ 32B & 0.530 \\
Gemma 3 27B & Qwen 3 32B & 0.503 \\
Gemma 3 12B & Aya Exp.\ 8B & 0.485 \\
\midrule
\multicolumn{3}{l}{\textit{Lowest agreement}} \\
Gemma 3 4B & Gemma 3 270M & $-$0.150 \\
Qwen 3 8B & Gemma 3 270M & $-$0.226 \\
Aya 8B & Gemma 3 270M & $-$0.231 \\
Qwen 3 1.7B & Gemma 3 270M & $-$0.267 \\
Gemma 3 1B & Gemma 3 270M & $-$0.277 \\
\bottomrule
\end{tabular}
\caption{Top-5 and bottom-5 pairwise Spearman $\rho$ of item-level correctness on CAPR-ja. Mean $\rho$: 0.165.}
\label{tab:cross-agree-ja}
\end{table}

\begin{table*}[!h]
\centering
\small
\begin{tabular}{rlrrrr}
\toprule
\textbf{ID} & \textbf{Cluster} & \textbf{Items} & \textbf{Avg (\%)} & \textbf{Best} & \textbf{Worst} \\
\midrule
20 & Greed \& Corruption            &  20 & 65.7 & 100.0 & 20.0 \\
18 & Prudent Moderation             &  80 & 64.6 &  83.8 & 22.5 \\
17 & Cost \& Worth                  &  40 & 64.3 &  87.5 & 17.5 \\
12 & Effort \& Resilience           &  90 & 64.0 &  85.6 & 27.8 \\
22 & Futile Efforts                 &  30 & 63.1 &  90.0 & 16.7 \\
 5 & Social Conduct \& Limits       & 100 & 60.9 &  82.0 & 24.0 \\
 6 & Growth \& Stagnation           &  50 & 60.3 &  92.0 & 22.0 \\
14 & Avoiding Danger                &  49 & 60.1 &  77.6 & 28.6 \\
 0 & Virtues \& Vices               &  59 & 57.0 &  84.7 & 18.6 \\
 3 & Interpersonal Bonds            &  88 & 56.3 &  76.1 & 27.3 \\
23 & Generosity \& Reciprocity      &  49 & 55.4 &  77.6 & 18.4 \\
 1 & Grounded Pragmatism            & 100 & 53.6 &  79.0 & 31.0 \\
19 & Patience \& Virtue             &  80 & 51.9 &  65.0 & 25.0 \\
 8 & Exploiting Opportunities       &  29 & 51.2 &  75.9 & 24.1 \\
 9 & Teamwork \& Unity              &  20 & 49.3 &  65.0 & 20.0 \\
21 & Hypocrisy Exposed              &  20 & 47.9 &  65.0 & 15.0 \\
13 & Unrealistic Hope               &  40 & 47.1 &  82.5 & 25.0 \\
 2 & Responsibility \& Consequence  &  39 & 46.9 &  61.5 & 17.9 \\
15 & Inevitable Consequences        &  89 & 45.4 &  60.7 & 16.9 \\
16 & Age \& Heritage                &  50 & 42.3 &  56.0 & 26.0 \\
 4 & Trust \& Betrayal              &  60 & 40.4 &  60.0 & 25.0 \\
 7 & Family Legacy \& Bonds         &  20 & 32.1 &  60.0 & 10.0 \\
10 & Pretentious Facades            &  20 & 26.8 &  65.0 &  0.0 \\
\bottomrule
\end{tabular}
\caption{Per-cluster accuracy on CAPR-ar, averaged across 14 models, sorted by descending accuracy. Best/Worst = highest/lowest single-model accuracy within the cluster.}
\label{tab:cluster-ar}
\end{table*}

\begin{table*}[!htbp]
\centering
\small
\begin{tabular}{rlrrrr}
\toprule
\textbf{ID} & \textbf{Cluster} & \textbf{Items} & \textbf{Avg (\%)} & \textbf{Best} & \textbf{Worst} \\
\midrule
16 & Generosity \& Gratitude             &  70 & 33.9 & 50.0 & 14.3 \\
 4 & Proper Timing \& Method             &  50 & 33.4 & 52.0 & 20.0 \\
 2 & Futile Efforts                      &  50 & 32.9 & 60.0 & 10.0 \\
11 & Greed \& Selfishness                &  29 & 32.8 & 62.1 & 10.3 \\
 8 & Honesty \& Deception                &  30 & 32.1 & 43.3 & 16.7 \\
 7 & Duty \& Responsibility              &  30 & 31.4 & 50.0 & 13.3 \\
12 & Humility \& Pride                   &  10 & 30.7 & 70.0 & 10.0 \\
 5 & Economic Inequality                 &  58 & 30.3 & 43.1 & 15.5 \\
14 & Reciprocity \& Value                &  40 & 29.3 & 47.5 & 17.5 \\
17 & Secrecy \& Betrayal                 &  70 & 28.8 & 44.3 & 22.9 \\
 1 & Foolishness \& Consequences         &  70 & 28.5 & 50.0 & 20.0 \\
10 & Inevitable Consequences             &  20 & 28.2 & 45.0 & 15.0 \\
 9 & Adversity \& Resilience             &  50 & 28.0 & 50.0 & 18.0 \\
 0 & Wisdom \& Moderation                &  40 & 27.1 & 40.0 & 12.5 \\
 6 & Self Knowledge \& Independence      &  40 & 26.8 & 60.0 & 10.0 \\
 3 & Risk \& Precaution                  &  39 & 26.7 & 56.4 & 12.8 \\
15 & Family Ties                         &  10 & 25.6 & 70.0 &  9.1 \\
19 & Strategic Planning                  &  10 & 20.0 & 40.0 &  0.0 \\
\bottomrule
\end{tabular}
\caption{Per-cluster accuracy on CAPR-am, averaged across 14 models.}
\label{tab:cluster-am}
\end{table*}

\begin{table*}[!htbp]
\centering
\small
\begin{tabular}{rlrrrr}
\toprule
\textbf{ID} & \textbf{Cluster} & \textbf{Items} & \textbf{Avg (\%)} & \textbf{Best} & \textbf{Worst} \\
\midrule
7 & Kindness \& Generosity         & 10 & 82.1 & 100.0 & 10.0 \\
4 & Patience \& Perseverance       & 10 & 80.0 & 100.0 & 30.0 \\
5 & Strategic Action \& Time       & 30 & 72.6 &  93.3 & 20.0 \\
1 & Truth \& Perception            & 10 & 65.7 & 100.0 &  0.0 \\
3 & Justice \& Equality            & 10 & 65.0 & 100.0 & 10.0 \\
6 & Boldness vs Caution            & 20 & 60.0 &  95.0 & 10.0 \\
0 & Perspective \& Influence       & 10 & 54.3 & 100.0 & 10.0 \\
\bottomrule
\end{tabular}
\caption{Per-cluster accuracy on CAPR-ja, averaged across 14 models.}
\label{tab:cluster-ja}
\end{table*}

\begin{table*}[h!]
\centering
\small
\begin{tabular}{l rr rr rr}
\toprule
 & \multicolumn{2}{c}{\textbf{CAPR-ar (1,222)}} & \multicolumn{2}{c}{\textbf{CAPR-am (716)}} & \multicolumn{2}{c}{\textbf{CAPR-ja (100)}} \\
\cmidrule(lr){2-3} \cmidrule(lr){4-5} \cmidrule(lr){6-7}
\textbf{Models correct} & \textbf{Items} & \textbf{\%} & \textbf{Items} & \textbf{\%} & \textbf{Items} & \textbf{\%} \\
\midrule
0--1    & 107 &  8.8 &  74 & 10.3 &  0 &  0.0 \\
2--3    & 118 &  9.7 & 219 & 30.6 &  3 &  3.0 \\
4--5    & 141 & 11.5 & 231 & 32.3 &  6 &  6.0 \\
6--7    & 162 & 13.3 & 131 & 18.3 &  9 &  9.0 \\
8--9    & 244 & 20.0 &  55 &  7.7 & 21 & 21.0 \\
10--11  & 234 & 19.1 &   6 &  0.8 & 34 & 34.0 \\
12--13  & 215 & 17.6 &   0 &  0.0 & 27 & 27.0 \\
14      &   1 &  0.1 &   0 &  0.0 &  0 &  0.0 \\
\bottomrule
\end{tabular}
\caption{Item difficulty distribution across all three benchmarks (14 models). Each row shows how many test items are answered correctly by the given number of models. Arabic is centered on 8--9 (well-calibrated), Amharic is skewed toward 2--5 (near-chance), and Japanese toward 10--13 (high accuracy).}
\label{tab:difficulty}
\end{table*}

\end{document}